\documentclass[sigconf]{acmart}

\AtBeginDocument{%
  \providecommand\BibTeX{{%
    \normalfont B\kern-0.5em{\scshape i\kern-0.25em b}\kern-0.8em\TeX}}}

\copyrightyear{2021}
\acmYear{2021}
\setcopyright{acmlicensed}\acmConference[CIKM '21]{Proceedings of the 30th ACM International Conference on Information and Knowledge Management}{November 1--5, 2021}{Virtual Event, QLD, Australia}
\acmBooktitle{Proceedings of the 30th ACM International Conference on Information and Knowledge Management (CIKM '21), November 1--5, 2021, Virtual Event, QLD, Australia}
\acmPrice{15.00}
\acmDOI{10.1145/3459637.3482290}
\acmISBN{978-1-4503-8446-9/21/11}
\begin{document}
\fancyhead{}
\title{Query-Variant Advertisement Text Generation \\
with Association Knowledge}


\author{Siyu Duan}
\affiliation{%
\institution{Research Center for Digital Humanities}
\institution{\& Department of Information Management,}
  \institution{Peking University}
  \city{Beijing}
  \country{China}}
\email{duansiyu@pku.edu.cn}

\author{Wei Li}
\affiliation{%
  \institution{School of Information Science,}
  \institution{Beijing Language and Culture University}
  \city{Beijing}
  \country{China}}
\email{liweitj47@pku.edu.cn}

\author{Jing Cai}
\author{Yancheng He}
\affiliation{%
  \institution{Platform and Content Group,}
  \institution{Tencent}
  \city{Shenzhen}
  \state{Guangzhou}
  \country{China}}
\email{{samscai,collinhe}@tencent.com}

\author{Yunfang Wu}
\author{Xu Sun}
\affiliation{%
  \institution{MOE Key Lab of Computational Linguistics,}
  \institution {School of EECS,}
  \institution{Peking University}
  \city{Beijing}
  \country{China}}
\email{{wuyf,xusun}@pku.edu.cn}

% \author{Yunfang Wu}
% \affiliation{%
%   \institution{MOE Key Lab of Computational Linguistics,}
%   \institution {School of EECS,}
%   \institution{Peking University}
%   \city{Beijing}
% \email{wuyf@pku.edu.cn}
\begin{abstract}
Online advertising is an important revenue source for many IT companies. In the search advertising scenario, advertisement text that meets the need of the search query would be more attractive to the user. However, the manual creation of query-variant advertisement texts for massive items is expensive. Traditional text generation methods tend to focus on the general searching needs with high frequency while ignoring the diverse personalized searching needs with low frequency. In this paper, we propose the query-variant advertisement text generation task that aims to generate candidate advertisement texts for different web search queries with various needs based on queries and item keywords. To solve the problem of ignoring low-frequency needs, we propose a dynamic association mechanism to expand the receptive field based on external knowledge, which can obtain associated words to be added to the input. These associated words can serve as bridges to transfer the ability of the model from the familiar high-frequency words to the unfamiliar low-frequency words. With association, the model can make use of various personalized needs in queries and generate query-variant advertisement texts. Both automatic and human evaluations show that our model can generate more attractive advertisement text than baselines. 
\footnote{Code and data set are available at https://github.com/CissyDuan/Query\_Variant\_AD.}
\end{abstract}
%https://www.overleaf.com/project/5f4c6bdf6bf2fe000130d38dƒ


\begin{CCSXML}
<ccs2012>
<concept>
<concept_id>10010147.10010178.10010179.10010182</concept_id>
<concept_desc>Computing methodologies~Natural language generation</concept_desc>
<concept_significance>500</concept_significance>
</concept>
<concept>
<concept_id>10010147.10010178.10010187</concept_id>
<concept_desc>Computing methodologies~Knowledge representation and reasoning</concept_desc>
<concept_significance>500</concept_significance>
</concept>
</ccs2012>
\end{CCSXML}

\ccsdesc[500]{Computing methodologies~Natural language generation}

\ccsdesc[500]{Computing methodologies~Knowledge representation and reasoning}
\keywords{advertisement; text generation; knowledge; graph; search}

\maketitle

\section{Introduction}
% The importance of the task
\begin{figure}
    \centering
    \includegraphics[width=0.8\linewidth]{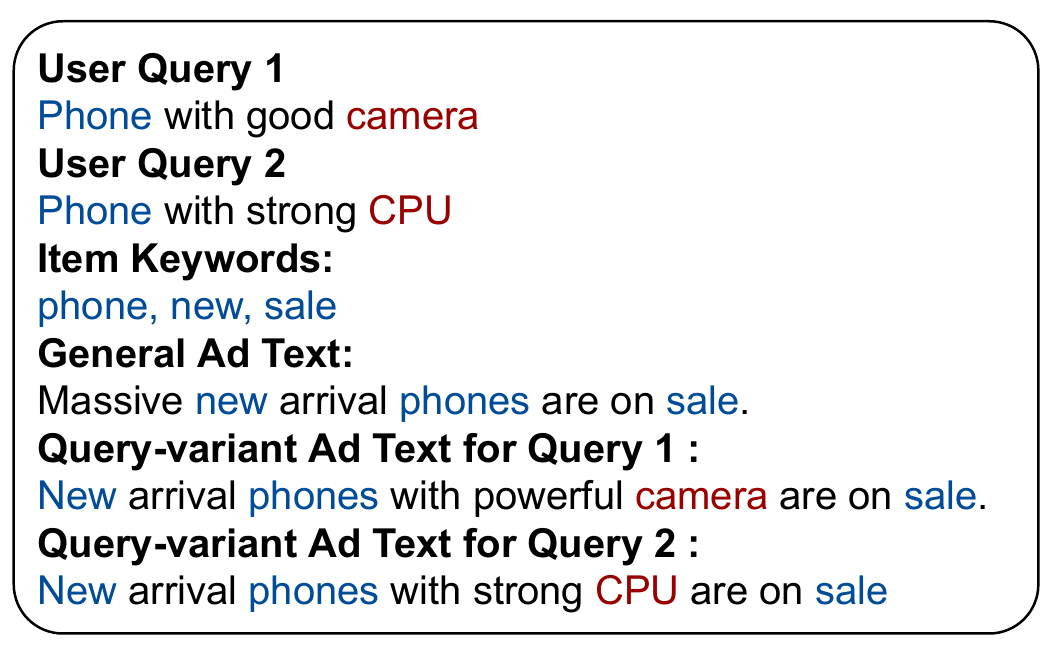}
    \caption{An example of the proposed query-variant advertisement text generation task. \textbf{User Queries} are natural language words given by users with various needs. \textbf{Item Keywords} are provided by ad sponsors, contains the core information of an item that usually reflects general needs.}
    \label{fig:example}
\end{figure}

The revenue of many Internet companies relies on advertising. In a typical online web search advertising system, each time a user inputs a search query in natural language words, the advertising system would respond with a list of advertisements of different items. However, one item involves different aspects. One predefined general purposed advertisement text can not address the focus of diverse personalized queries. If the advertising system can provide advertisement texts that are closer to user needs considering the search queries, the ads will be more attractive to the users.

An example of query variant advertisement text is given in Figure \ref{fig:example}. In this example, query 1 and query 2 pay attention to the camera and CPU performance separately, which are different selling points for the same item. The general purposed advertisement text fails to address these personalized needs for the two queries, while the ideal query-variant advertisement texts would emphasize the diverse needs that users are interested in. 
Although being a favourable feature for the advertising system, the manual creation of all the advertisement texts targeting different queries with different needs for massive items is too expensive.
%任务提出,为什么要用q/k/ad作为数据

\begin{figure}[t]
    \centering
    \includegraphics[width=0.8\linewidth]{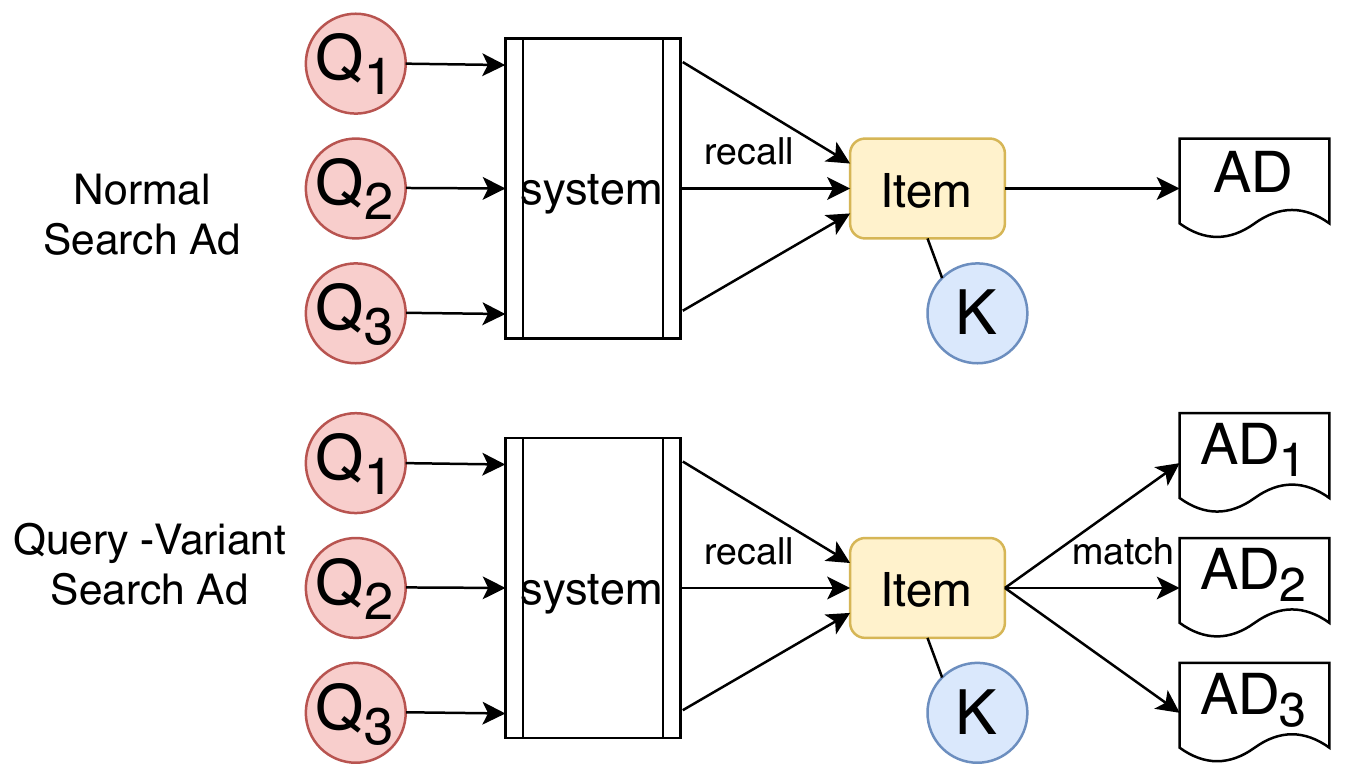}
    \caption{The process of search ad system. ``K'' is item keywords, ``Q'' is user query. Query-variant search ad system would respond the user query with more focused ad text for each item, which arouses a demand for the proposed query-variant ad text generation.} %Due to the loose coupling problem of user queries, traditional text generation method tends to treat the personalized needs in queries as noise and ignore them.}
    \label{fig:search-ad}
\end{figure}

In this paper, we propose the query-variant advertisement text generation task, which aims to automatically generate different advertisement texts considering the needs of various queries. 
This task serves the query-variant search ad system. An illustration of query-variant search ad system is shown in Figure \ref{fig:search-ad}. 
%, for which our proposed query-variant advertisement text generation task is designed.
%难点
Although query-variant advertisement text generation is a very practicable task, it faces a serious challenge that traditional text generation models tend to ignore the personalized needs with low frequency, while overemphasizing general purpose needs with high frequency.

\begin{figure}[t]
    \centering
    \includegraphics[width=0.6\linewidth]{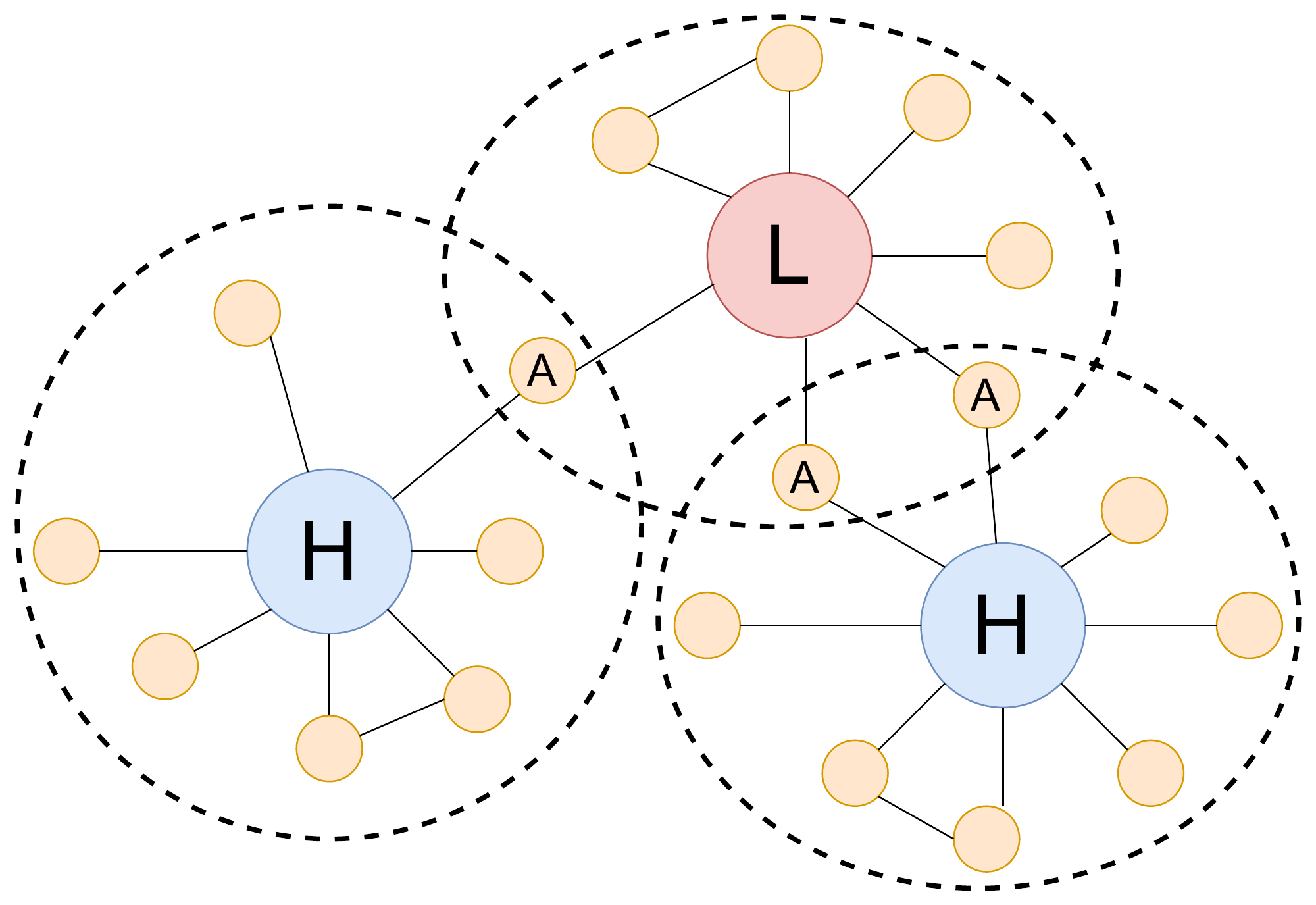}
    \caption{Association and Activation. The red circle``L'' represents the low-frequency words encountered, and the blue circles ``H'' represents the high-frequency words that the model dealt with in the training. 
    The orange circles represent the associated words, which expands the receptive field of the model. 
    The orange circles with ``A'' represent the shared words that are activated when being associated by our model. These associated words allow the modeling ability to transfer from familiar high-frequency words to unfamiliar low-frequency words.}
    \label{fig:activate}
\end{figure}

To improve the performance of the model on the low-frequency needs, we propose a dynamic association method to expand the receptive field of the model based on external knowledge. 
The mechanism of association is shown in Figure \ref{fig:activate}. The proposed association module can explore and select the suitable associated words based on both global structural information from external knowledge and local semantic information within the query-item pair. These selected associated words, which serve as the bridge between high-frequency keywords and low-frequency keywords, will be added to the original input words. When encountering low-frequency words, the associated words(the nodes with ``A'' inside in Figure \ref{fig:activate}) shared with high-frequency words could be activated. The ability of the model to deal with the familiar high-frequency words can be passed to the encountered low-frequency words through these activated associated words. With association, the model can make better use of the personalized needs contained in a specific query. 

For example, the cellphone brand ``Apple'' is much more popular than ``Blackberry'', and they share the associated word ``cellphone''. When encountering the low-frequency word ``Blackberry'', the ability to deal with the high-frequency word ``Apple'' can be transferred to ``Blackberry'' through ``cellphone''.

To test the effectiveness of our method, we collect over \textit{2,140,000} queries with the retrieved advertisement texts. We also build an association knowledge word graph based on \textit{17,000,000} advertisement texts.
%In addition to the automatic evaluation, we conduct human evaluations on the attractiveness, informativeness, and fluency of the generated advertisement text. 
Extensive experiments and human evaluations show that our model outperforms all the baselines, including the retrieved general purposed human written advertisement text.

\begin{figure*}
    \centering
    \includegraphics[width=0.8\textwidth]{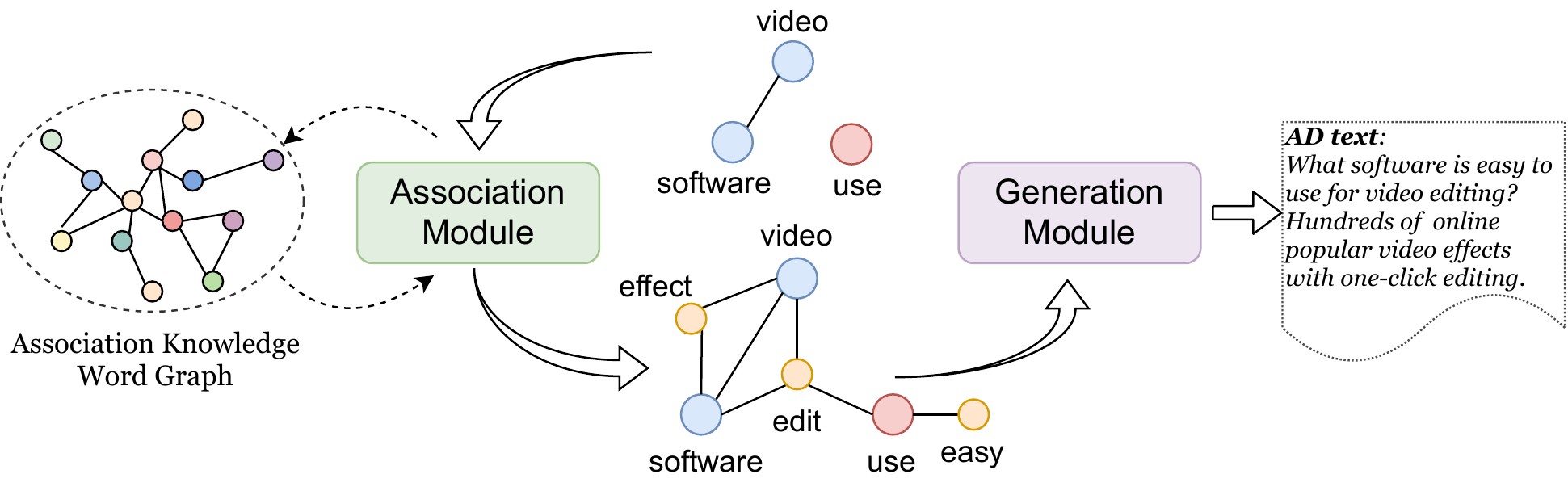}
    \caption{A sketch of the proposed model. {\color{blue}Blue circles} are item keywords, {\color{red}red circles} are query words, {\color{orange}orange circles} are the associated words retrieved by the Association module from the Association Knowledge Word Graph. The Generation module is designed to generate advertisement text based on the extended sub-graph.}
    \label{fig:model-full}
\end{figure*}

% contributions
Our contributions are summarized as follows:
\begin{itemize}
    %\item We propose the query-variant advertisement text generation task that aims to generate advertisement texts that satisfy the needs for various queries. We publish the collected dataset for further research.
    \item We apply query-variant advertisement text generation task to search advertisement scenario. This task aims to generate different advertisement texts that satisfy the needs for various queries.
    \item We propose a query-variant advertisement text generation model with a novel association method to transfer the modeling ability from familiar high-frequency words to unfamiliar low-frequency words, so as to better deal with the diverse and low-frequency personalized needs.
    \item We conduct extensive experiments and analyses. Both automatic and human evaluation results show that our model can generate attractive advertisement texts for various queries.
\end{itemize}

\begin{table*}[th]
\centering
\caption{Statistics of the dataset. q/k indicates the number of user queries and item keywords.}
%AKWG indicates the \textbf{A}ssociation \textbf{K}nowledge \textbf{W}ord \textbf{G}raph we build.}
\label{tab:data}
\resizebox{.8\textwidth}{!}{
\begin{tabular}{|l|ccccccc|}
\hline
Data & Train &Dev ( q / k )&Test ( q / k )&Max Len& Vocab & AKWG-node & AKWG-edge\\
\hline
Number & 2,126,938 & 4,516 / 1,458 & 4,174 / 1,321 & 20 & 50,000 & 49,460 & 869,371\\ 
\hline
\end{tabular}}
\end{table*}

\section{Task and Data}
In this section, we introduce our proposed task and the collection of experimental data.
\subsection{Task}
Before introducing our task formulation, we first describe the definitions of some phrases in this paper. \\
\textbf{Query and Item Keywords:}
Queries are the natural language search content of users in the website. Item keywords are isolated words containing the core information of the item provided by advertisement sponsors.\\
\textbf{General and Personalized Needs}: 
In online search advertising, for multiple user queries matching the same item, we treat the overlap words of different queries as general needs and the others as personalized needs. General needs are often included in item keywords, while personalized needs are distributed in different search queries. 
%In online searching advertising, we define general needs as the high-frequency keywords, which often appear in different search queries that match the same item. In contrast, personalized needs are other low-frequency keywords in the search queries. 
\\
\textbf{High-Frequency and Low-Frequency Words}:
We calculated the average Document Frequency and vocabulary size of general needs words and personalized needs words in the test set with multiple queries. The average Document Frequency of general needs words is 1.88 times that of personalized needs words, while the vocabulary size of personalized needs words is 1.52 times that of general needs words. It reveals that general needs words are high-frequency and centralized words, while personalized needs words are low-frequency and diverse words. 
 Concretely, in the example shown in Figure \ref{fig:example}, ``phone, new, sale'' rank at 44, 149, 761 respectively out of 50,000 words, while ``camera, CPU'' rank at 2070, 4921 respectively. The general needs (phone, new, sale) appear much more often than the personalized needs (camera, CPU). 
 %The general needs are high-frequency and centralized, while personalized needs are low-frequency and diverse.
\\
\textbf{Task Formulation}:\\
Given the user query $q$, the item keywords $k$ and the original advertisement text $t$ of the item. 
%The user queries contain different personalized needs, while item keywords contain the core information of the item that reflect the general needs for the item. 
The objective of the task is to generate advertisement texts $ads$ that are coherent to the item information while being closer to the focus of the given user queries(query-variant ads). 
\subsection{Data Collection}
We collect the data from the search advertising system of the Tencent QQ Browser. Some statistics of the dataset are given in Table \ref{tab:data}.\\
\textbf{Query-Ad Pair Data}\\
We use the query-ad pairs that are retrieved by online search advertising.
Each pair of data consists of three parts: query \textit{q} from the user, human written advertisement text \textit{t} retrieved by the system and keywords \textit{k} of this item. 
The keywords are provided by the advertisement sponsors. The maximum length of the advertisement text is limited to 20 words. The number of item keywords and the length of user query in each case are limited to 10 words. This length constraint setting balances the coverage of data (95\%) and calculation consumption. Furthermore, to ensure that the item keywords can reflect core item information, we set the number of the item keywords to be at least 3. 
%Our proposed method does not use user queries during the training phase, but some baseline models require it. In order to make the number of training cases the same, we use advertisement as the training unit. 
%In the training set we match the query with the maximum exposures for each item.
%In the test and validation set, we match multiple different queries for the same item to test its ability to generate query-variant ads.
There are 2.14 million different advertisement texts in our data, of which 2.12 million are used as the training set. For the test set and the validation set, the advertisement text that matches at least 2 queries in the system are selected from 10,000 advertisement texts. The test set is built in this way to evaluate the ability to satisfy the needs of different user queries.\\
%, and each advertisement is matched with different queries.
\textbf{Association Knowledge Data}\\
%The 50,000 words with top word frequency are used as the vocabulary. 
We propose to construct the association knowledge word graph based on the co-occurrence information between words in the original advertisement text. The co-occurrence word graph extracted from the advertising text corpus with the same distribution brings the historical co-occurrence information of words in advertising text. We collect 17 million advertisement texts to build an \textbf{A}ssociation \textbf{K}nowledge \textbf{W}ord \textbf{G}raph (AKWG). Formally, we use point-wise mutual information (PMI) to calculate the correlation between words. % $w_i$ and $w_j$:
%$$PMI(w_i, w_j) = log\frac{p(w_i, w_j)}{p(w_i) p(w_j)}$$
%where $p(w_i, w_j)$ is the probability of $w_i$ and $w_j$ appearing in the same %advertisement, $p(w_i)$ is the probability of $w_i$ appearing in the advertisement. 
In the $AKWG$ $(\mathcal{N}, \mathcal{E})$, the nodes $\mathcal{N}$ are the words, and an edge $e(i, j) \in \mathcal{E}$ is built between $w_i$ and $w_j$ when the $PMI(w_i, w_j)$ reaches a threshold $\xi$. In our construction setting, edges with PMI scores bigger than 8 are built in the $AKWG$, and the maximum number of neighbors for a word is limited to 20. Edge weight is normalized by dividing PMI by 8. 

\section{Approach}
%First, we give the task formulation, then we briefly describe the forward process of our model. Then we explain the modules in detail one by one and finally describe the training procedure.
In this section, we formally introduce our proposed query-variant advertisement text generation method. Our model mainly consists of two modules, the \textbf{Association module} and the \textbf{Generation module}. A sketch of our model is given in Figure \ref{fig:model-full}.

The association module is designed to expand the receptive field of the model based on external knowledge, which can help improve the ability of the model in dealing with diverse personalized needs.

%so that the discrete words (nodes) input is transformed into a linked word graph (group) with denser input space, resulting in more robust input distribution. 
%This helps the transition between training and inference more smoothing when queries are added during inference.
%input changes between two stages.
%information gap between the user query and item keywords. 
The generation module is designed to generate advertisement text based on the original input and the associated knowledge retrieved by the association module.
%Therefore, we treat the two types of input separately and make use of the graph structure information.
%The user query is added as input together with item keywords during inference.

For each item, the forward process is as follows:
\begin{enumerate}
    \item \textbf{Sub-graph Construction}: We first construct the sub-graph of input words $\mathcal{A}$ based on the item keywords $k$ and user queries $q$, which are connected by referring to the \textit{AKWG}.
    %Words from the user queries are added in this step during the inference stage.
    \item \textbf{Knowledge Association}: We associate external knowledge to $\mathcal{A}$ by referring to \textit{AKWG} with the association module and obtain the extended sub-graph $\mathcal{A}'$. This step follows a coarse-to-fine paradigm by considering both statistical and semantic information.
    \item \textbf{Ads Generation}: Finally, we generate the advertisement text based on the extended sub-graph $\mathcal{A}'$.% with the generation module. 
\end{enumerate}

\subsection{Sub-graph Construction} \label{sec:graph_construction}
%Our association mechanism is based on graph structure. The original input words and the associated words are connected through a graph structure, which reflects the path of association.
%Therefore, we choose to model the input discrete words into a sub-graph. 
%The sub-graph connects isolated keywords that reflect several different aspects of the item. 
%For a specific case, the relationship and status of its keywords are different. 
User queries are natural language text with many expressions, while item keywords are discrete words. To combine the two types of input, we choose to construct user query and discrete keywords into a sub-graph through the graph structure in $AKWG$, which provides historical information reflecting the co-occurrence between words in user query and item keywords.
%Further more, in proposed association mechanism, the graph structure reflects the path of our association.
%This entire sub-graph involves both the semantic information of a specific case and the structure information indicating the historical word co-occurrence.

Assume the bag-of-words containing item keywords $k$ and words in the user query $q$ as $\alpha$, we use the edges in \textit{AKWG}$(\mathcal{N}, \mathcal{E})$ to link the original input words. Formally, if $w_i \in \alpha$ , $w_j \in \alpha$ and $e(i, j) \in \mathcal{E}$, we build an edge between $w_i$ and $w_j$ in the sub-graph $\mathcal{A}$ .%, which provides the global correlation information among keywords $k$ of the item.

%Note that words of the user query $q$ are added to the sub-graph $\mathcal{A}$ during inference.
%Note that item keywords and user queries have different coupling degrees. Therefore, in the training phase, the construction of the sub-graph is based only on item keywords. In the inference phase, the construction of the sub-graph uses both item keywords and user queries. More details about the input setting strategy are introduced in section \ref{sec:Training and Inference}.
\begin{figure}
    \centering
    \includegraphics[width=0.98\columnwidth]{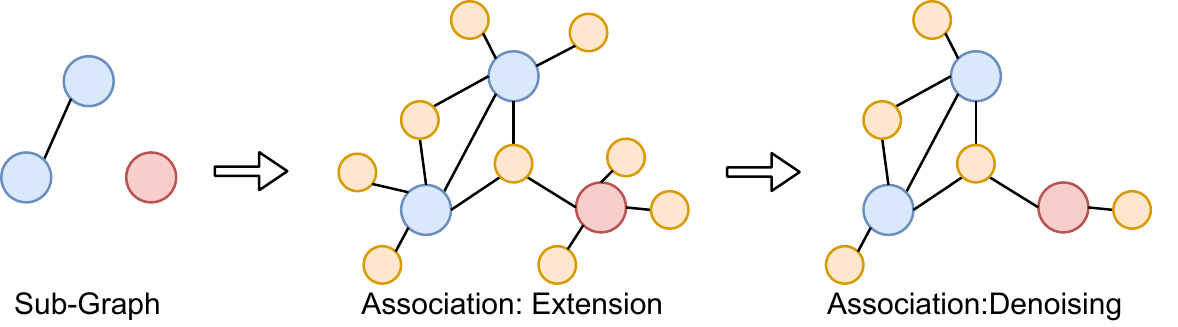}
    \caption{A sketch of association process. {\color{blue}Blue circles} are item keywords, {\color{red}red circles} are query words, {\color{orange}orange circles} are the associated words retrieved by the Association Module from the AKWG.}
    \label{fig:association}
\end{figure}

\subsection{Association Module}
%To deal with the large variance in the input data distribution, we propose a word-level association method by referring to external knowledge. 
The Association Module is designed to expand the receptive field of the model based on external knowledge, which can help improve the ability of the model in dealing with diverse personalized needs. It can retrieve the relevant external knowledge and combine it into the input words. The input words can be linked and expanded according to the graph structure in the $AKWG$. When the model encounters diverse and low-frequency input words, the associated words shared with some high-frequency words could be activated. The activated associated words can be regarded as a bridge to help the model transfer its ability in dealing with familiar high-frequency words to unfamiliar low-frequency words. In this way, the ability of the model to deal with low-frequency words can benefit from it. Furthermore, only applying statistical information would bring too much noise. Therefore, we propose to take advantage of the local semantic information within each case to select appropriate associated words.
To expand the receptive field and select the expanded information, we design two steps in the association module, \textbf{extension} and \textbf{denoising}. An illustration is shown in Figure \ref{fig:association}.

\noindent \textbf{Association: Extension}\\
We take the one-hop neighboring nodes of the input words($k$,$q$) in  $AKWG$ as the candidate nodes to be added to the sub-graph $\mathcal{A}$. The edges in $AKWG$ reflect the historical co-occurrence relation between words, which can help the model make reasonable associations. These candidate words that frequently co-occur with the keywords in the advertisement text data have similar semantic information with original input words, which allows the candidate words to be relevant. The candidate nodes introduced here can be noisy and needed to be filtered before being added to the graph.

\noindent \textbf{Association: Denoising}\\
In sub-graph extension, we roughly select candidate neighboring nodes by retrieving all the one-hop neighboring nodes of the input words in \textit{AKWG}. This rough selection procedure only makes use of the historical co-occurrence information and does not consider the semantic connection of different input words in each case, which means these candidate additional words can be noisy.
Therefore, to combine the semantic information in each case, we propose a reinforcement learning (RL) based method to score (select) the suitable candidates based on the semantic information and the graph structure within the sub-graph $\mathcal{A}$ of input words. 

We propose to use a \textbf{graph encoder} and a \textbf{score predictor} to score the candidates. The graph encoder is designed for encoding the sub-graph $\mathcal{A}$ into dense vectors, which contains the semantic information and structure information of the nodes. The score predictor is to predict the probability of the candidate nodes from \textit{AKWG} to be extended into the sub-graph $\mathcal{A}$ based on the graph representation learned by the graph encoder and the \textit{AKWG} edges. 
%为什么要用图

\noindent \textbf{Graph Encoder:} We propose a Gated Graph convolutional neural networks (GatedGCN) with gated attentive pooling as the graph encoder, which is calculated as follows:
\begin{align}
    \Tilde{H}^{l} &= GCN(H^{l-1}, adj) \\
    H^l &= LSTM(\Tilde{H}^l, H^{l-1}) \\
    \Tilde{g}^l &= AttnPooling(H^l) \\
    g^l &= LSTM(g^{l-1}, \Tilde{g}^l)
\end{align}
where $H^l$ is the hidden states of the $l$-th layer in the graph, $adj$ is the normalized adjacency matrix following the setting of GCN \citep{kipf2016semi}, $g$ is the global representation of the graph learned with attentive pooling in \citep{DBLP:journals/corr/LiTBZ15}. Attentive pooling allows the model to get a graph representation with calculated importance weights.
%The initial hidden states $H^0$ are initialized with the embeddings of the keywords in the graph $\mathcal{A}$.

This encoder applies GCN to aggregate the neighboring information and uses the gate mechanism in LSTM \citep{hochreiter1997long} to decide which part of the aggregated information should be transmitted into the next layer in the update of both node and global representation. 
GCN can combine semantic information and topological structure information to encode features. The gate mechanism of LSTM can select out important information between the two layers of GCN.
%This design takes advantage of the graph structure by applying GCN while releasing the over-smoothing problem with LSTM, which is often faced by GCN.

\noindent \textbf{Score Predictor:} After we get the global graph representations, we score the candidates with the \textbf{score predictor}:
\begin{equation}
    score = w([g^L; v_c])    \label{eqn:score_predictor}
\end{equation}
where $g^L$ is the global representation of the last layer, $v_c$ is the embedding of the candidate word $w_c$, $[;]$ means concatenation, and $w\in R^h$ is a learnable parameter matrix. The candidates with top $\phi$ scores are added to the original sub-graph $\mathcal{A}$, which form the new sub-graph $\mathcal{A}'$. 

\subsection{Generation Module}
After association, the original input words and the associated words are connected through the graph structure. We treat the original input words and the associated words as a kind of heterogeneous graph to make use of the associated words without damaging the original input information.
The generation module is designed to generate advertisement text based on the extended sub-graph $\mathcal{A}'$ obtained in the association. This module is developed based on Transformer \citep{vaswani2017attention}. Although Transformer has proved to be successful in modeling linear structure data, it is not suitable for the graph structure information in $\mathcal{A}'$.
%, which provides global co-occurrence information among words.

The graph structure of $AKWG$ contains prior information of massive advertising text, which can help our text generation task. To make use of the graph structure that links the input words, we propose to add a GatedGCN layer used in the association module before the Transformer encoder. The decoder follows the same architecture of the original Transformer. 
% \begin{equation}
%      H^1 = GatedGCN(X) 
% \end{equation}
To distinguish the role between the original input words and the associated knowledge words, we add type embedding $emb_t$ to the word embedding $emb_w$, $x=emb_w + emb_t$, which is similar to the positional encoding. In this way, the original input words and associated words can be seen as one kind of heterogeneous graph. The model can make use of the associated words without damaging the original input information.

\subsection{Training and Inference}
\label{sec:Training and Inference}
In the search advertisement system shown in figure \ref{fig:search-ad}, the ad of the same item can be recalled by multiple queries focusing on different aspects of the item. 
%This is because that the item keywords and user queries have different coupling degrees with the advertisement text. 
Item keywords usually indicate the general type of needs for the item, which is tightly coupled with the item. On the contrary, different user queries contain different personalized needs that are loosely coupled with the item. 
\begin{figure}[t]
    \centering
    \includegraphics[width=0.85\linewidth]{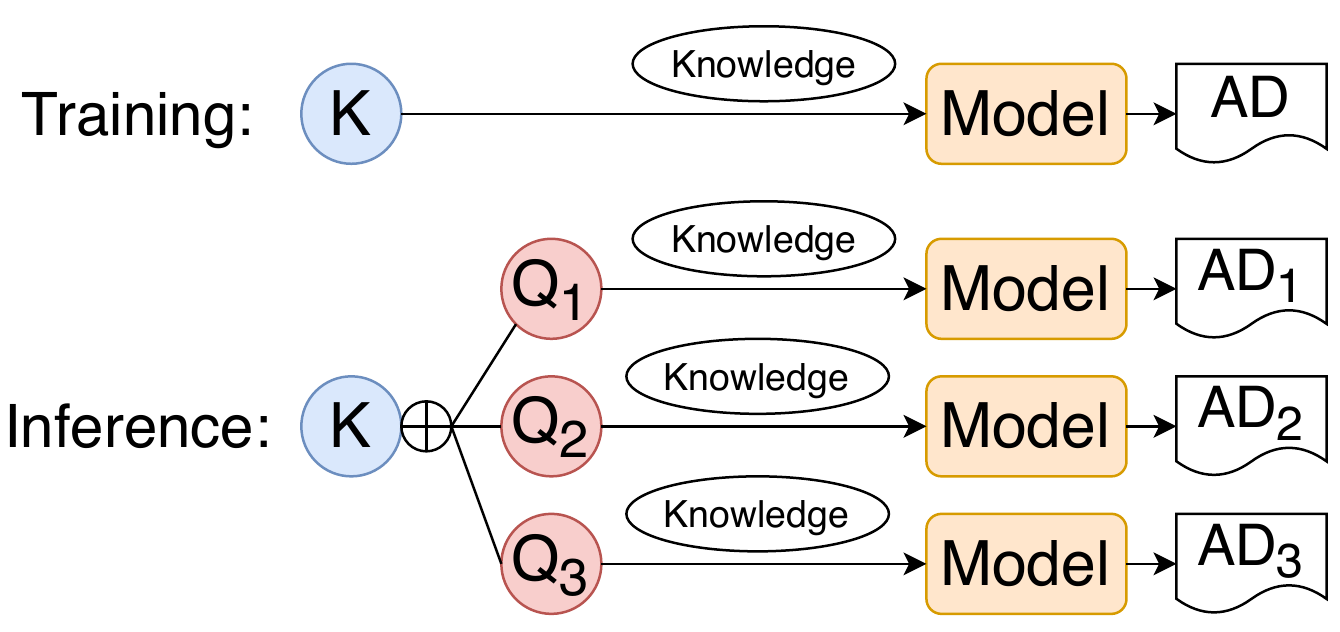}
    \caption{Different input settings between training and inference. ``K'' is item keywords, ``Q'' is user query. In the training phase, item keywords are used as input to fit the ad text. In the inference phase, item keywords and query are both used as input to generate query-variant advertisement text.}
    \label{fig:train-infer}
\end{figure}
If simply treating both keywords and queries as the input, the model tends to treat the loosely coupled parts of queries as noise and ignore them. As a consequence, the model would ignore the personalized needs and fail to generate query-variant advertisement text.
%Due to the different coupling degrees between item keywords and query, we need to prevent the loosely coupled queries from bringing noise to the model during the training phase. 

Therefore, we use different input setting strategies in training and inference. An illustration is shown in Figure \ref{fig:train-infer}. 
In the training phase, we use item keywords as input and advertisement text as generation target.
In the inference phase, words in the query are treated as discrete words like item keywords. Item keywords and query are both used as input to generate query-variant advertisement text. 
%For details, we split the query of the natural sentence into word bags and removed the stop words, and then merged with the item keywords to construct the sub-graph.
In this way, the model in the training phase will not be confused by the multiple loosely-coupled queries. It can learn the ability to generate general purpose advertisement text from tightly-coupled item keywords during training, and generate advertisement text for each query in the inference.
%When we treat a query as discrete words like keywords, and together with keywords as input during inference, the advertisement text generated by the model is for a specific query. It contains both the core information of the item and query preference. 
%Because the model learns the ability to generate advertisement text from discrete keywords during training, %when the query is treated as discrete words like keywords, together with item keywords as input during inference, it can generate advertisements that contain both the core information of the item and query preferences.
%when the queries are treated as discrete words like keywords, it can provide the essential query preferences for the model to generate query-variant advertisements.
%The input distribution variance caused by the different input settings in the training and inference can be controlled by the stable-variance method provided by our association module.
%Our goal of training is to make the advertisement generated from item keywords close to the original advertisement.
%To make the model able to generate coherent advertisement when adding query information as input during inference, we propose an RL based method to train the association module that can bridge the gap between the training stage and the inference stage caused by the absence of query in the supervised training. 
%Since the gradient can not be passed between the two modules,
%We propose a three-stage training process that applies supervised training to facilitate RL training (a.k.a teacher forcing). 

\begin{table*}[t]
\centering
\caption{Automatic evaluation results. \textit{Original} means the human written advertisement text retrieved by the advertising system. We show the change of \textit{Recall(q)} and \textit{Recall(q+k)} compared with the Original ads in parentheses. In \textit{Train} and \textit{Test} columns we show the input settings. \textit{k} is Item Keywords, \textit{q} is User Query. Our model beats all the baseline models in all the metrics except Dist-1. Although the diversity promoting model Transformer+itf gets a higher Dist-1 score, its Recall (q) is very low, which means the generated text is not targeted on the query.}
\resizebox{.82\textwidth}{!}{
\begin{tabular}{|l|clllll|cc|}
\hline
Model & BLEU & Recall(k) & Recall(q) & Recall(q+k) & Dist-1 & Dist-2 & Train & Test\\
\hline
Original&——&95.45 &65.54&74.51&9.28&21.73&——&——\\ \hline
Pointer Network&21.58&84.43 &75.91 (+10.37)&77.70 (+3.19)&11.11&36.65&k&k+q\\
Transformer&23.65&90.77 &88.06 (+22.52)&87.66 (+13.15)&13.07&38.85&k&k+q\\
GPMM&20.69&90.10 &63.45 (-2.09)&71.04 (-3.47)&7.96&20.20&k+q&k+q\\
ExpansionNet&21.40&89.68 &66.26 (+0.72)&71.87 (-2.64)&8.88&25.80&k+q&k+q\\
Transformer+itf&10.79&89.77 &64.40 (-1.14)&71.13 (-3.38)&\textbf{14.84}&25.81&k&k\\
\hline
Proposal&\textbf{27.12}&\textbf{92.36} &\textbf{91.03} (+\textbf{25.49})&\textbf{90.28} (+\textbf{15.77})&13.53&\textbf{41.36}&k&k+q\\
\hline
\end{tabular}}

\label{tab:baseline}
\end{table*}
\noindent \textbf{Training:} We propose a three-stage training process that facilitates RL training with supervised training.

In the \textbf{first} stage, we omit the score predictor in the forward pass. The extended keywords are randomly selected from the candidates (words connected with $\mathcal{A}$ in the \textit{AKWG}). This \textbf{first} supervised training process gives the model the clue of the extended graph with new words from the \textit{AKWG}, which can be used to roughly train the parameters (e.g. embeddings) of both the association module and the generation module. The generation module is trained for 15 epoch with cross-entropy weighted by the sentence length.
%$$
%    Loss =-\frac{1}{n}\times\sum_{j=1}^n{t_{j}log(p_j)}
%$$

In the \textbf{second} stage, we use policy gradient \citep{williams1992simple} based RL to train the association module. This stage aims to train the module to select the associated words by referring to the supervised objective of the task. %We use the model trained in the first stage to get the reward that guides the training of the association module. 
%which uses semantic information to filter out the appropriate association words to achieve the effect of denoising external knowledge.
The reward is calculated by comparing the advertisement text generated by the model in the first stage with the original human-written one. Formally, the reward is calculated as follows:
%$$
\begin{equation}
    %Reward = -log(tanh(-\frac{1}{n}\times\sum_{j=1}^n{t_jlog(p_j)}))
    Reward = 1-tanh(-\frac{1}{n}\times\sum_{j=1}^n{t_jlog(p_j)})
\end{equation}
%$$
where $t$ is the gold label, $p$ is the prediction probability.
%The probability that the associated word to be sampled is taken as the probability in the RL phase. 
For different inputs, the value of the sampling probability is affected by the total number of sampled words, thus affecting the RL loss.
Therefore the probability for each sample is normalized by the total number of words. %with which it is associated:
%$$
%    Prob = \sigma(\frac{a}{\phi }\times\sum_{i=1}^\phi{Prob_i}-1)
%$$
%where $\phi$ is the predefined associated word number, $a$ is the total associated candidate word number.
In this stage, the embedding of the generated model learned in the first stage is used as the embedding of the association module.
%, and the embedding parameters of this part are fixed. The random selection strategy in first stage is used as the baseline of reinforcement learning. 
Fine-tuning for reinforcement learning costs 5 epochs. 

In the \textbf{third} stage, we also apply supervised training after the reinforcement learning process when the parameters of the association module are settled.
%This stage can help the generation module adapted to the fixed association module. The loss function at this stage is the normal cross-entropy. 
%In order to make the association module able to filter the associated knowledge and let the generation module be fully adapted to the knowledge enhanced sub-graph $\mathcal{A}'$, we carry out a two-step training on the generation module in the third stage training. In the first step, the association module samples candidate neighbors according to the calculated importance score as the sampling probability to obtain the associated words. In the second step, the candidates with top-10 importance scores calculated by the association module are used to supplement the graph. In these two steps of the third stage, we train the generation model with normal entropy loss for 15 epochs. 

\noindent \textbf{Inference:} In inference, we treat the query as discrete words like keywords. These words, together with item keywords, are used as input during inference. We split the query of the natural sentence into a bag of words and removed the stop words, and merged with the item keywords. The sub-graph is constructed with the merged bag of words, which forms the basis of association and generation.
%Both item keywords and words in query are used in sub-graph construction, which is later used to retrieve external knowledge and generate advertisements. 
%The sub-graph constructed in this way is used in subsequent association module and generation module to generate query-variant ad text.

\section{Experiment}
In this section, we give the experiment results. Extensive analysis is conducted to demonstrate the effectiveness of our model.

\subsection{Setting}
 The embedding size is 128. The hidden size is 256. 
 For a fair comparison, the generation module of our model is obtained by replacing the first layer in the Transformer encoder with the mentioned GatedGCN, which has parameters of the same scale as Transformer. Both the GatedGCN and the Transformer encoder have two layers, while the decoder has three layers. We use Adam optimizer \citep{kingma2014adam} to train the model.
 Considering the memory consumption and model performance, we set the value of $\phi$ to 10 in our experiment.

%Although the queries are not involved in the training phase in our model, they are needed by some baselines. To make all the models (including our model and the baselines) trained with equal times, we train the models adjusted by the advertisement number.

\subsection{Baseline}
In this work, we compare our model with five strong baselines in addition to the original human-written one. The numbers of parameters are all of the comparable scales to our model.

\noindent \textbf{General End-to-End Baselines:}\\
To compare the modeling capabilities of the proposed models, we introduced two end-to-end text generation models with different architectures.
\begin{itemize}
    \item
    \textbf{Pointer network} \citep{see2017get}: A Seq2Seq model with copy mechanism. 
    The attention mechanism is used to calculate the additional probability of the input words to be combined with the original output probability. This baseline is applied because the advertisement text usually shares many words in the item keywords and user query.

    \item
    \textbf{Transformer} \citep{vaswani2017attention}: Powerful text generation model. Our generation module is a graph suitable improvement over it.
    %Both the encoder and decoder use three-layer multi-head self-attention to get the context representation. The encoder and decoder are connected by cross-attention.
\end{itemize}

\noindent \textbf{Personalized Text Generation Baselines:}\\
The task proposed in this paper is close to the personalized text generation task that uses search queries as personalized information, so we introduce two personalized text generation baselines.
\begin{itemize}
    \item
    \textbf{GPMM} (Generative Profile Memory Network) \citep{DBLP:conf/acl/KielaWZDUS18}: A Seq2Seq based method aiming at personalized text generation. We adapt the model to our task by treating each word of user query as individual memory representations in the memory network.
    %During the decoding, the hidden state of decoder do attention on the memory in each decoding step.
    \item
    \textbf{ExpansionNet} \citep{DBLP:conf/acl/NiM18}: Another Seq2Seq based method aiming at personalized text generation, which calculates two attention scores between decoder hidden state and the word embedding in the user query. 
    %One for combining weighted query embedding with decoder hidden state and the other for giving additional probability for words in user query when generating predicted word.
\end{itemize}

\noindent \textbf{Diversity Promoting Baseline:}\\
The proposed task hopes to obtain diversified texts corresponding to different search queries, so we add a pure diversity promoting baseline.
\begin{itemize}
    \item
    \textbf{Transformer + itf} \citep{nakamura2018another}: A diversity-promoting method that weights the loss function by \textit{inverse token frequency}. The network is the same as Transformer.
\end{itemize}
The Seq2Seq models in GPMM, ExpansionNet and Pointer Network are all LSTM-based \citep{hochreiter1997long} Seq2Seq model with attention mechanism \citep{luong2015effective}. Both encoder and decoder have three layers.

\subsection{Automatic Evaluation}
% evaluation method

\noindent \textbf{Metrics}:\\
%We choose \textbf{BLEU} (comprehensive)\cite{papineni2002bleu} , \textbf{recall rate},  the automatic evaluation metrics. 
we choose three widely applied automatic metrics to evaluate the quality of the generated text regarding the expression ability, query coverage, and diversity. 
%The expression ability indicates the comprehensive ability of the model to generate text. If the model can fit general-purpose advertisements well when using keywords as input, then when query is added as input, the model can also generate smooth advertisements corresponding to specific queries.
%The query coverage indicates the reflection of the generated advertisement text to query information. This measures whether our model is effective for query variant advertisement generation task.
%And the diversity indicates whether the model has a variety of language expressions. Tedious, boring advertisements are meaningless to our task.

\textbf{BLEU} \cite{papineni2002bleu}: BLEU is calculated between the original human written advertisement text and the one generated under the same input setting during training. This metric is to measure the expression ability of the model from item keywords to advertisement text.
If the model can fit general-purpose advertisement text well when using keywords as input, then when the query is added as input, the model can also generate smooth advertisement text corresponding to specific query.
%Note that the ads generated for one item will only be measured once.
%In the baseline where user query is required as input during the training phase, for each item, we use the query that has the max exposures with the item as input when calculating BLEU.In other models, BLEU is calculated only the item keywords are used as input.This setting makes the model face the same data distribution as the training set when calculating BLEU and ads generated by each item will only be calculated once.

\textbf{Recall}: Recall(k), Recall(q), Recall(q+k) calculate the recall between the generated advertisement text and item keywords (k), user query (q), the union of item keywords and user query (q+k) respectively in inference. We also present the recall improvement over the original advertisement text in parentheses to measure the ability to absorb the query information. 
%In the test set, one item is matched with different queries. 
We calculate the recall by averaging over the results of different queries of the item, which reflects the ability to generate query-variant advertisement texts when facing different queries for the same item.
%This is also a sign of the ability to be adapted to different queries while retaining the core information of the item.

\textbf{Dist-1, Dist-2} \citep{DBLP:conf/naacl/LiGBGD16}: 
These two metrics count the number of distinct unigrams and bigrams divided by the total number of unigrams and bigrams in the generated text. These two metrics reflect informativeness and diversity.
%Larger number of vocabulary generally means better informativeness and diversity. 

% to measure the expression ability of the model, which also reflects the ability to reserve the key information out of the item. 

% \begin{table}[t]
%     \centering
%     \caption{Human evaluation of content coherence. We evaluate the proportion of generated ads coherent to item content. Transformer w/o query means that the transformer model does not use query in the inference phase. Compared with the baseline that is not disturbed by query, the acceptance rate is slightly lower but acceptable.}
%     \resizebox{.95\columnwidth}{!}{
%     \begin{tabular}{c|c}\hline
%         Model  &  coherence (p=0.43) \\ \hline
%         Transformer w/o query & 90.20 \\
%         Proposal & 85.80 \\ \hline
%     \end{tabular}}
%     \label{tab:coherence}
% \end{table}

% 说明三个问题：
% 1.原本的广告recall低，因此任务本身很有必要
% 2.线性模型不适合于此任务
% 3.生成好的广告需要额外的信息
\begin{table*}[th]
    \centering
    \caption{Human evaluation results. ``win'' indicates the ratio of the generated advertisement text that our proposed model is better than the counter model. ``win\textgreater lose'' means our proposal is better than the other one. The Pearson's r for each metric is shown in the header within the braces.% ``NoQuery'' means not using queries when doing inference.
    }
    \resizebox{.75\textwidth}{!}{
    \begin{tabular}{|l|ccc|ccc|ccc|}\hline
    Versus &
            \multicolumn{3}{c|}{Attractiveness (p=0.44)} &
            \multicolumn{3}{c|}{Informativeness (p=0.46)} & \multicolumn{3}{c|}{Fluency (p=0.22)} \\
            &  win & lose & tie & win & lose & tie & win & lose & tie \\ \hline
        Proposal-Original  & 57.27 & 8.47 & 34.27 & 60.53 & 21.73 & 17.73 & 25.40 & 19.53 & 55.07 \\
        Proposal-Transformer & 38.00 & 14.87 & 47.13 & 48.07 & 22.33 & 29.60 & 21.67 & 10.93 & 67.40\\
        Proposal-NoQuery& 58.67 & 4.13 & 37.20 & 69.07 & 9.07 & 21.87 & 21.73 & 18.93 & 59.33 \\
    \hline
    \end{tabular}}
    \label{tab:human_evaluation}
\end{table*}
\noindent \textbf{Result}: \\
In Table \ref{tab:baseline} we show the automatic evaluation results of our model compared with the baselines and the original advertisement text. 
%From the results, we can see that our model outperforms all the baseline models on all three metrics. 
We can observe that the Recall(q) of the original advertisement text is significantly lower than Recall(k), showing that the original coverage of query words is not satisfactory, which means that the general purposed advertisement text can not meet the diverse personalized needs of different queries focusing on different aspects of the item. This phenomenon testifies the necessity of our proposed task, which aims to generate query-variant advertisement text. 

From the results, we can also see that traditional personalized text generation baseline methods (GPMM, ExpansionNet) do not perform well. Due to the loose-coupling problem of item-queries that one item can be matched with different queries, the model tends to treat the queries as noise and ignore them in the training, which makes the model unable to generate various advertisement texts corresponding to different queries in the inference. As a result, these common methods that model query information in the training phase are not suitable for search advertisement text generation task. It also proves that our different input setting strategies in the training and inference phase are necessary.

Note that although the pure diversity promotion method (Transformer + itf) gets a higher ratio in Dist-1, the Recall(q) is too low to be applicable, which means the generated text ignores the query. This indicates that general-purpose diversity promotion method is not suitable for the proposed task where the generated advertisement texts should answer the needs of different user queries.

Compared with the Transformer and the Pointer Network that take the same input setting in training and inference with our model, our model achieves much better results for all metrics. 
This is because that the associated external knowledge not only brings more fruitful information but also enhances the ability of the model to deal with words with different exposure frequency, so that the model can better fit the inputs in the training phase, and achieve better generalization effects after adding query in the inference phase. 
The proposed model gains the highest Recall(q) improvement among different user queries, which means that our proposed method can be adapted to variant user queries and generate query-variant advertisement texts.

\begin{figure}
    \centering
    \includegraphics[width=0.99\linewidth]{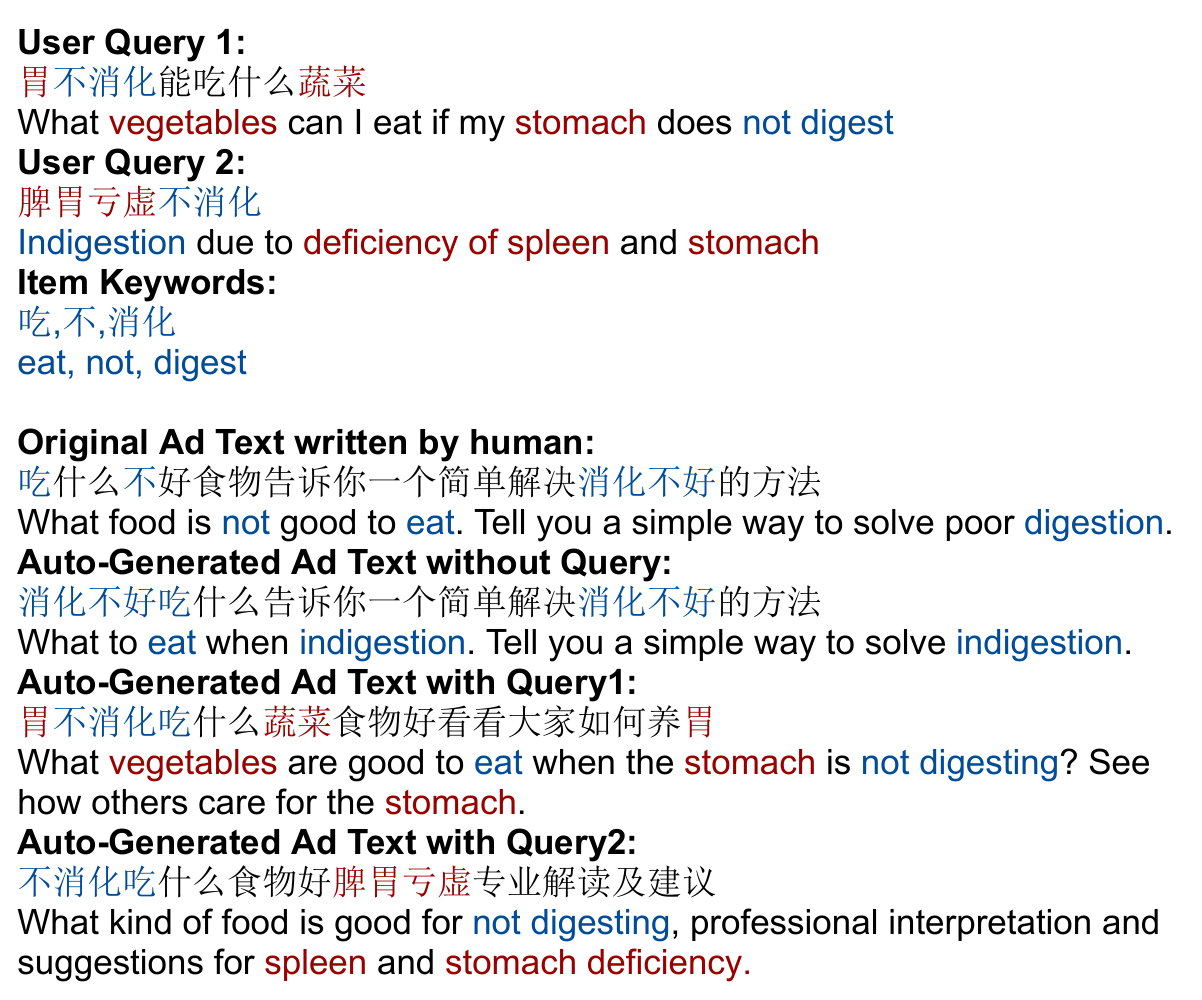}
    \caption{Concrete example of advertisement text generation. Blue words represent general needs information, which is usually included in the item keywords that match the item. Red words represent diverse personalized needs contained in different user queries.}
    \label{fig:case}
\end{figure}

\begin{table*}[t]
\centering
\caption{Ablation study on the effect of association module and graph structure in generation module. 
\textbf{-Association} means that the association module is removed. 
\textbf{-Graph Structure} means to replace GatedGCN in the generation module with normal Transformer encode layer.
%Our model is built on the basis of Transformer.
%\textbf{+GatedGCN} means to add GatedGCN to the generation module without association module.
%\textbf{+Association} means that only the association module is added, and the generation module is the original transformer.
%\textbf{+query (input)} means adding the query to the input during training. 
%We show the recall change in parentheses compared with the original human written one.
}\label{tab:gnn}
\resizebox{.7\textwidth}{!}{
\begin{tabular}{|l|clllll|cc|}
\hline
Model & BLEU & Recall(k) & Recall(q) & Recall(q+k) & Dist-1 & Dist-2\\
\hline
\textbf{Proposal}&\textbf{27.12}&\textbf{92.36} &\textbf{91.03 (+25.49)}&\textbf{90.28} (+\textbf{15.77})&\textbf{13.53}&\textbf{41.36}\\
-Association&23.91&89.39 &86.54 (+21.00)&86.16 (+11.65)&13.18&39.05\\
-Graph Structure&25.77&91.89&90.05 (+24.51)&89.97 (+15.16)&13.39&40.66\\
%-Association\&Graph&23.65&90.77 &88.06 (+22.52)&87.66 (+13.15)&13.07&38.85\\
%\hline

%\hline
% +query (input)&22.54&87.75 &74.06 (+8.52)&77.96 (+3.45)&11.15&32.04&k+q&k+q\\
\hline
\end{tabular}}
\end{table*}

\begin{table*}[t]
\centering
\caption{Effect of different retrieving methods. ``Random'' means randomly selecting $\phi$ neighbors from \textit{AKWG}. ``PMI'' means selecting neighbors with top-$\phi$ PMI scores. ``no-filter'' means selecting all the neighbors in \textit{AKWG}. ``Proposal'' means selecting with our proposed association module. 
%The changes shown in the parentheses are the recall compared with the original human written one.
}\label{tab:retrieve}
\resizebox{.65\textwidth}{!}{
\begin{tabular}{|l|clllll|}
\hline
Model & BLEU & Recall(k) & Recall(q) & Recall(q+k) & Dist-1 & Dist-2\\
\hline
random&24.52&91.60&89.52 (+23.98)&89.09 (+14.58)&13.48&41.13\\
pmi&24.77&91.65&89.66 (+24.12)&89.14 (+14.63)&13.39&40.46\\
no-filter&25.20&90.63&88.41 (+22.87)&87.82 (+13.31)&13.14&39.19\\
\hline
\textbf{Proposal}&\textbf{27.12}&\textbf{92.36} &\textbf{91.03} (+\textbf{25.49})&\textbf{90.28} (+\textbf{15.77})&\textbf{13.53}&\textbf{41.36}\\
\hline
\end{tabular}}
\end{table*}

\subsection{Human Evaluation}
\noindent \textbf{Evaluation Data}: \\
In the human evaluation, we focus on the cases where the default retrieved advertisement text does not meet the needs of the queries.
%We sample advertisements that are divergent from the queries as human evaluation data. 
%Because they are the cases where the given advertisements can not meet the need of the query.
If more than 50\% of the characters in the query without stop words do not appear in the advertisement text, we assume that there exists a major divergence between the query and the retrieved advertisement text, which are chosen as the candidates for human evaluation. After sensitive data violating privacy issues is automatically filtered out, we randomly sample 500 cases as human evaluation data and invite 3 human annotators to do the evaluation. Evaluators are employees hired by IT companies to review various online text, which means they have expertise in reviewing advertisement text.

\noindent \textbf{Evaluation Guideline}: \\
In human evaluation, the user query and two advertisement texts options are provided to the evaluators. The evaluators are asked to choose the better one out of the two optional advertisement texts in terms of three aspects:\\
\begin{itemize}
    \item
    \textbf{Attractiveness} is a comprehensive and the most important indicator, representing which advertisement text the users prefer given the query.\\
    \item
    \textbf{Informativeness} represents which advertisement text provides more useful information regarding the user query.\\
    \item
    \textbf{Fluency} represents the language smoothness of the advertisement text. 
\end{itemize}

%For each evaluation data pair, the experts are asked to choose the better advertisement of each indicator.
If the two advertisements are equally good (bad), they are given a tie. 
In the final statistical results, if win $>$ lose, it means that our proposed model beats the baseline.

% \subsubsection{Correlation}
%  among annotators \& automatic evaluation

%Because the result is to compete the result of one model versus the other, when ``win $>$ lose'', our model beats the other one.

\noindent \textbf{Result}: \\
\noindent In Table \ref{tab:human_evaluation} we show the human evaluation results. We compare our model with three baselines, namely, the original human written advertisement text, Transformer and $NoQuery$. $NoQuery$ means the result is generated by the proposed model without query during inference, which means that the query information is totally ignored. Results show that our proposed model beats all the baselines, especially in attractiveness and informativeness. The results on \textbf{attractiveness} and \textbf{informativeness} show good correlation among different evaluators, while rather low correlation on \textbf{fluency}. This is expected because the concept of being fluent is hard to define, especially in the advertisement text area.

We provide concrete advertisement text generation cases in Figure \ref{fig:case}. The advertisement generated by our method takes into account both general needs contained in item keywords and query preference, which is not available in general-purpose human written advertisement and advertisement generated without query.

\subsection{Ablation Study}
\noindent \textbf{Effect of Association and Graph}: \\
Table \ref{tab:gnn} shows the ablation study results on association and graph structure. 
The removal of \textbf{associative thinking} (-Association) causes the decrease of the BLEU score and the Recall (q+k) by 3.21 and 4.12. We think this is because the ability of the model to deal with different types of words is benefited from the association module. This allows the model to achieve better fitting results in training. In the inference phase, the model can better absorb the query information, which makes the model less likely to generate irrelevant advertisement text. 

As for the \textbf{graph structure} in the generation module, after replacing the GatedGCN with the normal Transformer encode layer(-Graph Structure) in the generation module, the BLEU score drop by 1.35. This shows that the graph structure with prior information can help to organize discrete input information into a fluent natural language sentence.

\noindent \textbf{Effect of Different Retrieval Methods}:\\
In Table \ref{tab:retrieve} we show the results of different retrieving methods. Among all the retrieving methods, our proposed method achieves the best on all the metrics. 

Although the ``no-filter'' method adds all the one-hop neighbors on $AKWG$ to the input, its performance on all metrics is still weaker than our proposed model. We think that this is because there is too much noise brought by adding all the neighbors in \textit{AKWG} to the input sub-graph. This testifies that the denoising procedure of the association module is necessary.

``PMI'' signifies the correlation of the candidate knowledge and gains higher recall improvement (+14.63) than ``random'' and ``no-filter''. However, the PMI score only concerns the global historical co-occurrence knowledge, while our method not only makes use of the PMI score via the edge weight in the GCN component in the association module but also considers the semantic information of the keywords encoded in the graph representations. This design helps our method achieve the best BLEU score (27.12) and the highest recall improvement (+15.77) among all the retrieval methods.
% test the effect of RL 
\begin{figure}
    \centering
    \includegraphics[width=0.95\linewidth]{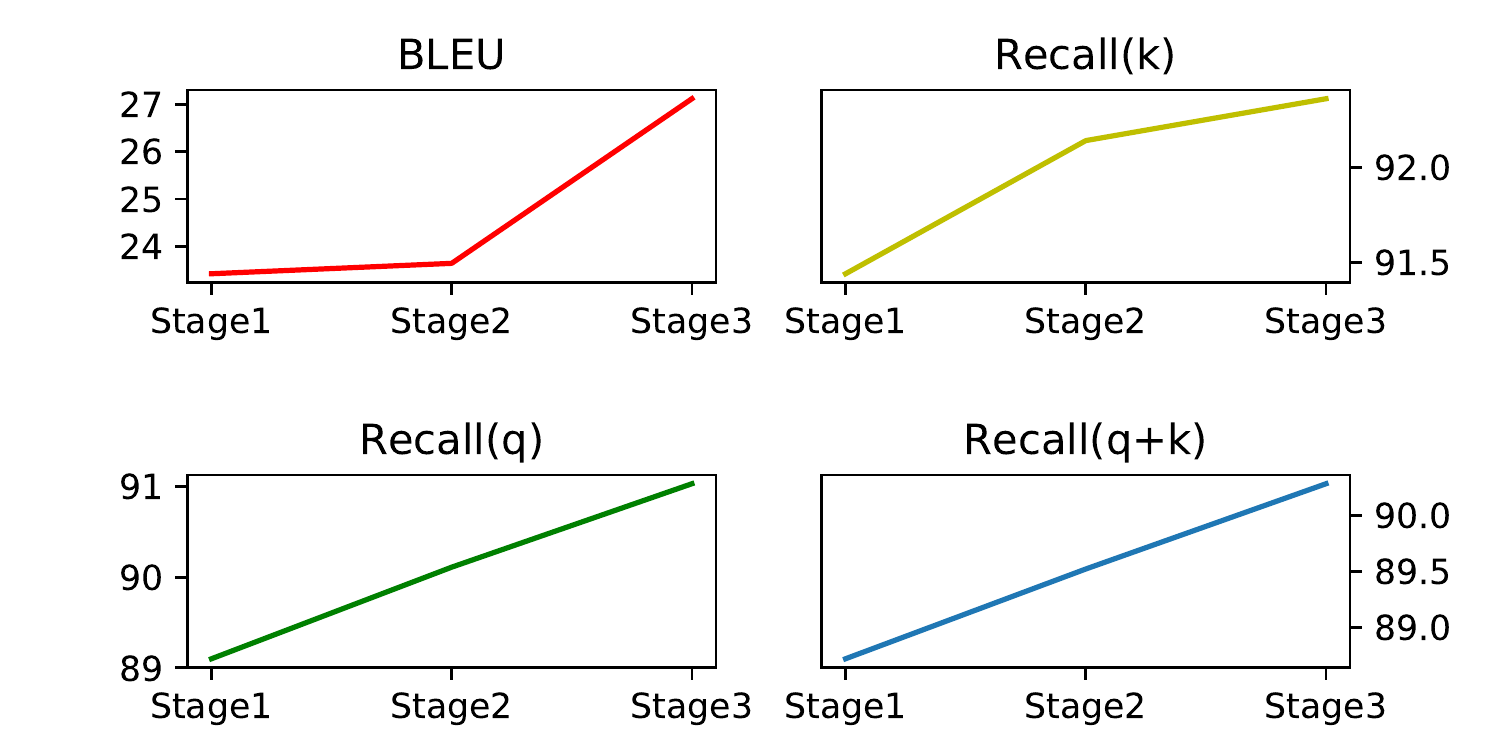}
    \caption{Results of different training stages. Stage 2 is RL based. Stage 1 and stage 3 are supervised.}
    \label{fig:RL}
\end{figure}

\noindent \textbf{Effect of RL}:\\
In Figure \ref{fig:RL} we show the performance trend during the RL training stages (dictated in section \ref{sec:Training and Inference}). From the figure, we can see that all three stages are necessary to train the model. In the reinforcement learning stage, the improvement proves that the association module can denoise the associated candidate words after reinforcement learning and obtain association words that dynamically consider the semantic information in the sub-graph of a specific case. With the help of the two supervised stages, RL training can indeed improve the performance of the model.

\section{Related Work}
\noindent \textbf{Product Information Generation:} Product text information (e.g., advertisement, product description) is widely displayed in various Internet scenarios. Previous researchers have conducted many studies for different application scenarios. 
\citet{wang2017statistical} propose a statistical framework that generates product descriptions from product attributes. 
\citep{DBLP:conf/kdd/ChenLZYZ019,10.1145/3397271.3401140} generate personalized product descriptions on e-commerce platforms. 
\citet{chan2019stick} utilize the entity information for fidelity-oriented product descriptions.
\citep{10.1145/3292500.3330754,10.1145/3308558.3313536} use user clicks feedback information to guide the generation of advertisement text. 
\citet{Wang_Tian_Qiu_Li_Lang_Si_Lan_2018} apply a multi-task learning approach for improving product title compression with user search log data. 
\citep{chan2020selection} propose a new task, i.e., how to generate a multi-product advertisement post.
The task we proposed is applied to search advertising, which is to generate advertisement text for a specific query. 
%Most methods for generating personalized advertisement text require online user information, which inevitably violates user privacy. 
Our method is to generate query-variant ad text offline, which are then recalled online. The existing methods cannot completely solve the problems in our task scenarios. 
%The method we propose does not use any user information in the training and inference phase.

%\citet{DBLP:conf/kdd/ChenLZYZ019} propose a knowledge based model to generate personalized product description. 
%\citet{10.1145/3397271.3401140} generate product descriptions on e-commerce platforms, and we explore this classic task from a novel perspective that allows the optimal output text to vary with ever-changing user preference.
%\citet{10.1145/3292500.3330754} Generating Better Search Engine Text Advertisements with Deep Reinforcement Learning
%\citet{10.1145/3308558.3313536} put forward a novel seq2seq model to generate social advertisements automatically, in which a quality-sensitive loss function is proposed based on user click behavior to differentiate training samples of varied qualities.

\noindent \textbf{Personalized and Diverse Text Generation:} Many previous works have been proposed to use personalized information for text generation. 
\citep{DBLP:conf/acl/LiGBSGD16,DBLP:conf/acl/KielaWZDUS18,chan2019modeling} make use of speaker information and capture characteristics of speakers in neural conversation. 
\citet{DBLP:conf/aaai/XingWWLHZM17} incorporate topic information into the Seq2Seq framework to generate informative and interesting responses for chatbots. 
\citet{DBLP:conf/acl/NiM18} aim at generating personalized reviews by expanding short phrases with aspect-level information. 
%\citet{DBLP:conf/acl/LiGBSGD16} present a persona-based model that captures characteristics of speakers in neural conversation.
%\citet{DBLP:conf/acl/KielaWZDUS18} make use of speaker profile information in chit-chat.
These works provide good views of personalized generation. However, there is one big obstacle making their methods not suitable for our proposed task, that is, there is no strict coupled personalized query-advertisement pair in our task. In our experiment, we compared several personalized text generation methods and revealed their inapplicability to the proposed task.
%Besides, in our task setting, we focus on the variant queries rather than user profiles.

There are also works focusing on generating general purposed diverse text.
With the idea of reducing the frequency of high-frequency text and increasing the frequency of low-frequency text, various methods \citep{li2015diversity,xu2018diversity,liu2018towards,nakamura2018another} have been used to increase the diversity of text generation. 
\citet{vijayakumar2016diverse} propose to diversify the outputs by optimizing diversity-augmented objective in beam search. 
Unlike the diversity-promoting works above, our proposed task aims to generate query-variant advertisement text that can diversify the advertisement text in the context of meeting the purpose of different query needs. Therefore, those methods that do not consider the queries are not applicable.
%这些personal的方法和我们的方法最大区别其实在于他们的personal信息和目标文本是一对一的，比如用户的类别信息，类别是固定的；一段目标文本只对应一个用户的个人资料描述；而我们的personal信息是文本的，基于内容的，对同一个目标文本，personal信息可以有不同的文本表达方法，是松耦合的。
%These works focus on using the user information rather than the specific content of queries. In consideration of privacy and industry needs, we focus on the query-variant part rather than user-variant.

%\subsection{Diverse Text Generation}
% \citet{li2015diversity} use MMI as the objective function in neural conversation models.
%\citet{xu2018diversity} use GAN to generate diverse text by assigning a low reward for repeated text and high reward for novel and fluent text. 
% \citet{liu2018towards} generate diverse text by reducing the cross-entropy loss of repeated text.  
% \citet{nakamura2018another} use ITF loss individually, which scales smaller loss for frequent token classes and larger loss for rare token classes. 
\noindent \textbf{External Knowledge:} External knowledge is an important information source for natural language understanding. 
Earlier attempts using the knowledge graph mainly focus on answering questions with a single clue retrieved from the knowledge graph \citep{hao-etal-2017-end,bordes-etal-2014-question,bordes2014open,dong2015question}. Although these works achieve big success, the applications are limited because of the idealized task setting. 
Other works propose to supplement the state-of-the-art model with external knowledge \citep{chen2016knowledge,chen2017neural,parthasarathi2018extending}. The knowledge usage in these works remains latent, and it is hard to tell the effect of the knowledge. 
\citet{zhou2018commonsense} use large-scale commonsense knowledge in conversation generation. 
\citet{mihaylov2018knowledgeable} integrate external knowledge in a cloze-style setting for the reading comprehension task. These works try to reason over the graph, which is a good way to use external knowledge. However, there is no pre-defined knowledge in our task. Therefore, we propose to first construct an association knowledge word graph based on the co-occurrence information in the advertisement text.

% \subsection{Graph-to-sequence Models}
% Owing to the ability to represent graph-structured data, graph-to-sequence based models have been applied to various tasks.
% 这句话介绍了graph based model的优点，其实也算是说明为什么我们要采用这个模型，属于我们和前人工作的共同点——利用graph model强大的表示能力
% \citet{zhao2018graphseq2seq} and \citet{beck2018graph} propose to use graph-based model to fuse the tree-structured syntactic and semantic information into machine translation respectively. These models only use graph-structured information to supplement the sequential data. \citet{xu2018sql,xu2018graph2seq} represents the SQL query as a directed graph in the SQL-to-Text task. Unlike their work, the keywords are not naturally organized as graphs in the advertisement generation task. %Therefore, we need to first find the relation between the attributes. 

% % 这里的脉络是开始大家只用现成的图结构数据，之后会改造成图，这里跟我们的方法是类似的，我们也需要先改造成图
% \citet{li2019coherent} proposes to organize the long articles into graph-structured data. Different from the abundant information contained in the long article, the information within the item keywords is quite limited. %Therefore, we not only need to organize the the unstructured data into graph structured data, but also need to enrich the graph with external knowledge.
% \citet{koncel2019text} generates coherent multi-sentence texts from the output of an information extraction (IE) system. The knowledge within the output of the IE system is self-contained, and do not involve external knowledge. 
%section{Task Formulation and Data Collection}

\section{Ethical Issues}
\noindent\textbf{Industrial Application:}
For the industrial application, the generated ad candidates will be further checked by human annotators before added to the online advertising system to prevent fake and misleading advertisement text from being shown to users. In the manual review phase, 85.80\% of generated ads are accepted. This acceptance rate can meet the efficiency and accuracy requirements of industrial applications. 
In the industrial application, item keywords and possible queries are provided by the advertising sponsor, which means that our method will not use user information in the training and inference phase.

\noindent\textbf{Dataset Open-Sourced:}
The data set contains search query content of website users and brand information of advertising sponsors. To protect the their privacy, the data set will be open-sourced after ID projection, which means that the mapping between word ID and word text will not be open-sourced. This approach allows peers to reproduce our automatic evaluation results and conduct computational experiments, but cannot get the original text information. Although this will affect the reproducibility of human evaluation, we believe that the protection of users and customers privacy is the highest priority.

%\section{Future Work}
%This work proposes a word-level association method to use historical knowledge to help text generation. 

\section{Conclusion and Future Work}
In this paper, we propose the query-variant advertisement text generation task that aims to generate candidate advertisement texts for different search queries with various needs. 
An association method based on external knowledge is proposed to improve the ability to deal with diverse personalized needs of search queries. Both automatic and human evaluation results show that our model can generate attractive and query-variant advertisement text. 
Our work provides a new perspective on the industrial application and research of search advertisement text generation.

Pre-training is an important method of using historical text knowledge. In future work, we want to explore how to incorporate pre-training into existing methods. 
Furthermore, our data and experiments are conducted on the Chinese data set. Because the quality of the Chinese open lexicon is not ideal, the lexicon we currently use only includes $AKWG$ constructed using historical advertising text. In the future, we will explore using lexicon from different information sources.
%The relatively small number of item keywords and the diverse expression of queries make the input distribution unstable, resulting in large data distribution variance. To deal with such problems, we propose a query-variant advertisement text generation model with a novel external knowledge association method. The word-level association on the external knowledge allows additional associated words to link and expand the input words through the graph structure. In this way, the input composed of discrete words (nodes) is transformed into a linked word graph (group) that has denser input space, obtaining a more robust input distribution. 
%To deal with the large data distribution variance, we propose a query-variant advertisement text generation model with a novel external knowledge association method.
%It helps to expand and denoise the input words through the graph structure. In this way, the discrete words (nodes) are transformed into a linked and expanded word graph (group) that has denser input space, resulting in a more robust input distribution.
%This input distribution stabilizing method allows the model to robustly cope with the input distributions with large variances, and generate advertisement text containing item information and query preferences.

%\bibliography{anthology,emnlp2020}
\bibliography{acmart.bib}
\bibliographystyle{ACM-Reference-Format}
\end{document}